\pdfoutput=1
\documentclass[12pt]{article}
\usepackage[utf8]{inputenc}
\usepackage[english]{babel}
\usepackage{csquotes}
\usepackage{fullpage}
\usepackage{fancyhdr}
\usepackage[pdftex]{graphicx}
\graphicspath{{./images/}}
\usepackage{setspace}
\usepackage{color}
\usepackage{float}
\usepackage{hyperref}
\usepackage{amsmath}
\usepackage{amsfonts}       
\usepackage{nicefrac}       
\usepackage{microtype}      
\usepackage{algorithm}
\usepackage{algpseudocode}
\usepackage{lipsum}
\usepackage{booktabs}
\usepackage[margin=1in]{geometry}
\usepackage{subcaption}
\newcommand{\ra}[1]{\renewcommand{\arraystretch}{#1}}
\usepackage{chngcntr} 

\usepackage{wrapfig}
\usepackage{hanging}

\usepackage[toc,page]{appendix}
\usepackage{rotating}
\usepackage{multirow}
\newcommand{\textcite}[1]{\citeauthor{#1}, \citeyear{#1}}
\providecommand{\keywords}[1]{\textit{Keywords:} #1}
 
\usepackage{natbib}
\title{OR-Gym: A Reinforcement Learning Library for Operations Research Problems}
\author{
	Christian D.~Hubbs,\thanks{Department of Chemical Engineering, Carnegie Mellon University, Pittsburgh, PA 15123} \\
	Hector D. Perez,\footnotemark[1] \\
	Owais Sarwar,\footnotemark[1]\\
	Nikolaos V. Sahinidis,\footnotemark[1] \\
	Ignacio E. Grossmann,\footnotemark[1] \\
	John M. Wassick\thanks{Dow Chemical, Digital Fulfillment Center, Midland, MI 48667}
}

\begin{document}

\maketitle

\begin{abstract}
Reinforcement learning (RL) has been widely applied to game-playing and surpassed the best human-level performance in many domains, yet there are few use-cases in industrial or commercial settings.
We introduce OR-Gym, an open-source library for developing reinforcement learning algorithms to address operations research problems.
In this paper, we apply reinforcement learning to the knapsack, multi-dimensional bin packing, multi-echelon supply chain, and multi-period asset allocation problems, and benchmark the RL solutions against MILP and heuristic models.
These problems are used in logistics, finance, engineering, and are common in many business operation settings.
We develop environments based on prototypical models in the literature and implement various optimization and heuristic models in order to benchmark the RL results.
By re-framing a series of classic optimization problems as RL tasks, we seek to provide a new tool for the operations research community, while also opening those in the RL community to many of the problems and challenges in the OR field.
\end{abstract}


\section{Introduction}

Reinforcement learning (RL) is a branch of machine learning that seeks to make a series of sequential decisions to maximize a reward \citep{Sutton2018}.
The technique has received widespread attention in game-playing, whereby RL approaches have beaten some of the world's best human players in domains such as Go and DOTA2 [\citet{Silver2017}, \citet{Berner2019a}]. 
There is a growing body of literature that is applying RL techniques to existing OR problems.
\citet{Kool2019} use the REINFORCE algorithm with attention layers to learn policies for the travelling salesman problem (TSP), vehicle routing problem (VRP), the orienteering problem (OP), and a prize collecting TSP variant.
\citet{Oroojlooyjadid2017} use a Deep Q-Network (DQN) to manage the levels in the beer game and achieve near optimal results.
\citet{Balaji2019} provided versions of online bin packing, news vendor, and vehicle routing problems as well as models for RL benchmarks.
\citet{Hubbs2020} use RL to schedule a single-stage chemical reactor under uncertain demand which outperforms various optimization models.
\citet{Martinez2011} approaches a flexible job shop scheduling problem with tabular Q-learning to outperform other algorithms.
\citet{Li2017} provides an overview of the progress and development of RL as well as a review of many different applications.

We seek to provide a standardized library for the research community who wish to explore RL applications by building on top of the preceding work and releasing OR-Gym, a single library that relies on the familiar OpenAI interface for RL \citep{Brockman2016}, but containing problems relevant to the operations research community.
To this end, we have incorporated the benchmarks in \citet{Balaji2019}, while extending the library to the KP, multi-period asset allocation, multi-echelon supply chain inventory management, and virtual machine assignment.
RL problems are formulated as Markov Decision Processes (MDP), meaning they are sequential decision making problems, often times probabilistic in nature, and rely on the current state of the system to capture all relevant information for determining future states. 
This framework is not widely used in the optimization community, so we make explicit our thought process as we reformulate many optimization problems to fit into the MDP mold without loss of generality.
Many current RL libraries, such as OpenAI Gym, have many interesting problems, but problems that are not directly relevant to industrial use.
Moreover, many of these problems (e.g. the Atari suite) lack the same type of structure as classic optimization problems, and thus are primarily amenable to model-free RL techniques, that is RL algorithms that learn with little to no prior knowledge of the dynamics of the environment they are operating in.
Bringing well-studied optimization problems to the RL community may encourage more integration of model-based and model-free methods to reduce sample complexity and provide better overall performance.
It is our goal that this work encourages further development and integration of RL into optimization and the OR community while also opening the RL community to many of the problems and challenges that the OR community has been wrestling with for decades.

The library itself along with example code to enable reproducibility can be found at \href{www.github.com/hubbs5/or-gym}{www.github.com/hubbs5/or-gym}.

\section{Background}

Included in the library are knapsack, bin packing, supply chain, travelling salesman, vehicle routing, news vendor, portfolio optimization, and traveling salesman problems.
We provide RL benchmarks using the Ray package for a selection of these problems, as well as heuristic and optimal solutions \citep{Moritz2018}.
Additionally, we discuss environmental design considerations for each of the selected problems.
Each of the environments we make available via the OR-Gym package is easily customizable via configuration dictionaries that can be passed to the environments upon initialization.
This enables the library to be leveraged to wider research communities in both operations research and reinforcement learning.

All mathematical programming models are solved with Gurobi 8.2 and Pyomo 5.6.2 to optimality on a a 2.9 GHz Intel i7-7820HQ CPU unless otherwise noted (\citet{Gurobi2018}, \citet{Hart2017}).

\subsection{Reinforcement Learning}

\begin{figure}
    \centering
    \includegraphics[width=0.5\textwidth]{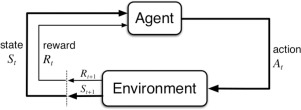}
    \caption{Diagram of a reinforcement learning system.}
    \label{fig:reinforcement_learning}
\end{figure}

Reinforcement learning (RL) is a machine learning approach that consists of an agent interacting with an environment over multiple time steps, indexed by $t$, to maximize the cumulative sum or rewards, $R_t$, the agent receives (see Figure \ref{fig:reinforcement_learning}, \citet{Sutton2018}).
The agent plays multiple episodes - Monte Carlo simulations of the environment - and at each time step within an episode, observes the current state ($S_t$) of the environment and takes an action according to a policy ($\pi$) that maps states to actions ($a_t$). The goal of RL is to learn a policy that obtains high rewards.
The problems are formulated as MDPs, thus RL can be viewed as a method for stochastic optimization \citep{Bellman1957}.

Deep reinforcement learning uses multi-layered neural networks to fit a policy function, with parameters $\theta$, that will map states to actions.
Here, we will use the Proximal Policy Optimization (PPO) algorithm \citep{Schulman2016} for our RL comparisons.
This is an actor-critic method, which consists of two networks, one to produce the actions at each time step (the \textit{actor}) and one to produce a prediction of the rewards at each time step (\textit{critic}).
The actor learns a probabilistic policy that produces a probability distribution over available actions.
This distribution is sampled from during training to encourage sufficient exploration of the state space.
The critic learns the value of each state, and the difference between the predicted value from the critic and the actual rewards received from the environment is used in the loss function ($\mathcal{L}(\theta)$) to update the parameters of the networks.
PPO limits the update of the network parameters by clipping the loss function (Equation \ref{eq:loss_function}) relative to the previous policy.
\begin{equation}
    \mathcal{L}(\theta) = \textrm{min} \big(r_t(\theta), 
    \textrm{clip} \big(r_t(\theta), 1 - \epsilon, 1 + \epsilon \big) \big) \hat{A}_t
    \label{eq:loss_function}
\end{equation}
where $r_t(\theta)$ is the probability ratio between the previous policy, $\pi_{k-1}(\theta)$, and the new policy $\pi_k(\theta)$ and $k$ denotes the updates to the policy parameters since initialization.
The function $\textrm{clip}$ enforces the constraint $1-\epsilon \leq r_t(\theta) \leq 1+\epsilon$ on the probability ratio.
$\epsilon$ is a hyperparameter that limits the update of the policy, such that the probability of outputs does not change more than $\pm \epsilon$ at each update.
$\hat{A}_t$ denotes the advantage estimation of the state, which is the sum of the discounted prediction errors over $T$ time steps given in Equation \ref{eq:advantage}.
\begin{equation}
    \hat{A}_t = \sum_{t=1}^T \gamma^{T-t+1}\delta_t
    \label{eq:advantage}
\end{equation}
where $\delta_t$ is the prediction error from the critic network at each time-step $t$ and $\gamma$ is the discount rate.
This modification to the loss function has shown more stable learning over other policy gradient methods across multiple environments \citep{Schulman2016}.

We rely on the implementation of the PPO algorithm found in the Ray package \citep{Moritz2018}.
All RL solutions use the same algorithm and a 3-layer fully-connected network with 128 hidden nodes at each layer, for both the actor and critic networks.
Although some hyperparameter tuning is inevitable, we sought to minimize our efforts in this regard in order to reduce over fitting our results.

\section{Knapsack}

The Knapsack Problem (KP) was first introduced by \citet{Mathews1896}, and a classic exposition of the problem can be found in \citet{Dantzig1957} where a hiker who is packing his bag for a hike is used as the motivating example.
KP is a combinatorial optimization problem that seeks to maximize the value of items contained in a knapsack subject to a weight limit.
Obvious applications of the KP include determining what cargo to load into a plane or truck to transport.
Other applications come from finance, we may imagine an investor with limited funds who is seeking to build a portfolio, or apply the framework to warehouse storage for retailers \citep{Ma2019}.

There are a few versions of the problem in the literature, the unbounded KP, bounded, multiple choice, multi-dimensional, quadratic, and online versions \citep{Kellerer2004}.
The problem has been well studied and is typically solved by dynamic programming approaches or via mathematical programming algorithms such as branch-and-bound.

We provide three versions of the knapsack problem, the Binary (or 0-1) Knapsack Problem (BinKP), Bounded Knapsack Problem (BKP), and the Online Knapsack Problem (OKP). 
The first two are deterministic problems where the complete set of items, weights, and values are known from the outset.
The OKP is stochastic; each item appears one at a time with a given probability and must be either accepted or rejected by the algorithm.
The online version is studied by \citet{Marchetti-Spaccamela1995}, who propose an approximation algorithm such that the expected difference between this algorithm and the optimal value is, on average, $\mathcal{O}(\textrm{log}^{3/2}n)$. \citet{Lueker1995} later improved this result with an algorithm that closes the gap to within $\mathcal{O}(\textrm{log}n)$ on average, using an on-line greedy algorithm.



\subsection{Binary (0-1) Knapsack}

The binary version of the knapsack problem (BinKP) can be formulated as an optimization problem as follows:
\begin{subequations}
\begin{gather}
    \max_{\mathbf{x}} z = \sum_{i=1}^n v_i x_i \\
    \textrm{s.t.} \; \sum_{i=1}^n w_i x_i \leq W \\ 
    x_i \in \rm \{0, 1\}    \quad i=1, \ldots, n
\end{gather}
\end{subequations}
where $x_i$ denotes an binary decision variable to include or exclude an item from the knapsack. The weights and values, $v_i$ and $w_i$ respectively, are positive, real numbers. The knapsack's total weight limit is denoted by $W$.
This model can be solved by pseudo-polynomial dynamic programming algorithms or, conveniently, as an integer programming problem using algorithms such as branch-and-bound to maximize the objective function.

\subsection{Bounded Knapsack}

The bounded knapsack problem (BKP) differs from the binary case in that $x_i$ becomes an integer decision variable ($x_i \in \rm \mathbb{Z}^{0+}$) and we introduce a new constraint based on the number of each item $i$ we have available:
\begin{equation}
    x_i \leq N_i    \quad i=1, \ldots, n
\end{equation}
where $N_i$ is the number of times the $i$th item can be selected.

\subsection{Online Knapsack}

A version of the Online Knapsack Problem (OKP) is found in \citet{Kong2019}.
This requires the algorithm to either accept or reject a given item that it is presented with.
There are a limited number of items that the algorithm can choose from and each is drawn randomly with a probability $p_i$, $i=1, \ldots, n$. 
After $M$ items have been drawn, the episode terminates leaving the knapsack with the items inside.
The goal here is the same as for the traditional knapsack problems, namely to maximize the value of the items in the knapsack while staying within the weight limit, although it is more challenging because of the uncertainty surrounding the particular items that will be available.

\subsection{Problem Formulation}

The RL model's state is defined as a concatenation of vectors: item value, item weight, number of items remaining, as well as the knapsack's current load and maximum capacity.
For the OKP cases, we provide the agent with the current item's weight and value, and the knapsack's load and maximum capacity.
The reward function for our RL algorithm will simply be the total value of all items  placed within the knapsack.
At each step, the RL algorithm must select one of the items to be placed into the knapsack, at which point the state is updated to reflect that selection and the value of the selection is returned as the reward to the agent.
The episode continues until the knapsack is full or no items fit.


The BinKP and BKP are solved as integer programs to optimality using Gurobi 8.2 \citep{Gurobi2018} and Pyomo 5.6.2 \citep{Hart2017}.
Additionally, we use a simple, greedy heuristic given in \citet{Dantzig1957}, which orders the items by value/weight ratio, and selects the next item that fits in this order.
If an item does not fit - or the item has already been selected $N$ times - the algorithm will continue through the list until all possibilities for packing while remaining within the weight constraint are exhausted, at which point the algorithm will terminate.
The OKP algorithm used is a greedy, online algorithm based on the \textit{TwoBins} algorithm given in \citet{Han2015}.
\begin{algorithm}
\caption{\footnotesize TwoBins Online Knapsack Algorithm}
\label{algo:okp}
\begin{algorithmic}[1]
\Require Choose $r \in \{0, 1\}$ with equal probability 
\For{Each item $i \in N$ in order}: 
    \If{$r=1$}: 
        \If{$w_i \leq W$}: 
            \State{Select $i$ for the knapsack} 
        \Else:
            \State{Reject $i$} 
        \EndIf
    \EndIf
    \If{$r=0$}: 
        \If{$\sum_i^n w_i \leq W$}: 
            \State{Reject $i$} 
        \Else:
            \State{Select $i$ for the knapsack} 
        \EndIf  
    \EndIf
\EndFor
\end{algorithmic}
\end{algorithm}
While IP solutions seek to solve the problem simultaneously, RL requires an MDP formulation which relies on sequential decision making.
In this case, the RL model must select successive items to place in the knapsack until the limit is reached.
This approach seems more akin to how a human would pack a bag, placing one in at a time until the knapsack is full.
The human knapsack-packer can always remove an item if she determines it does not fit or finds a better item, our RL system, however cannot: once an item is selected for inclusion, it remains in the knapsack.



\subsection{Results}

\begin{table}
\centering
\caption{RL versus heuristic and optimal solutions. All results are averaged over 100 episodes.}
\label{tab:kp_results}
\scalebox{0.75}{
\begin{tabular}{@{}clccc@{}}\toprule
Knapsack Version & Metric & RL & Heuristic & MILP \\
\midrule
      & Mean Rewards       & 1,072 & 1,368 & 1,419 \\
BinKP & Standard Deviation & 66.2  &     0 &     0 \\
      & Performance Ratio vs MILP & 1.32   & 1.04 & 1 \\
\midrule
    & Mean Rewards       & 2,434 & 2,627 & 2,696 \\
BKP & Standard Deviation &   207 &     0 &     0 \\
    & Performance Ratio vs MILP  & 1.11  &  1.03 & 1 \\
\midrule
 & Mean Rewards       & 309  &  262  &  531  \\
OKP & Standard Deviation & 54.2 &  104 &  54.6 \\
   & Performance Ratio vs MILP & 1.8 &  2.02 & 1 \\
\bottomrule
\end{tabular}
}
\end{table}

The knapsack problem and its variations have been widely studied by computer scientists and operations researchers. 
Thus, good heuristics exist for these problems, making it challenging for model-free methods to compete.
However, as shown in Table \ref{tab:kp_results}, the RL method comes close to the heuristics and optimal solutions in the deterministic, offline cases, and outperforms the heuristic in the stochastic case.
While RL does learn a good policy in the deterministic cases, the policy is probabilistic, thus it continues to exhibit some variance once trained.

In the case of the Online Knapsack, the RL model learns a policy that is close to the theoretical performance ratio for the \textit{TwoBins} algorithm.
The \textit{TwoBins} algorithm under performs its theoretical level ($\approx 1.7$) because each episode has a step limit, which causes the \textit{TwoBins} algorithm to pack too little before the limit is reached, bringing down its mean performance.

\begin{figure}
    \centering
    \includegraphics[width=0.8\textwidth]{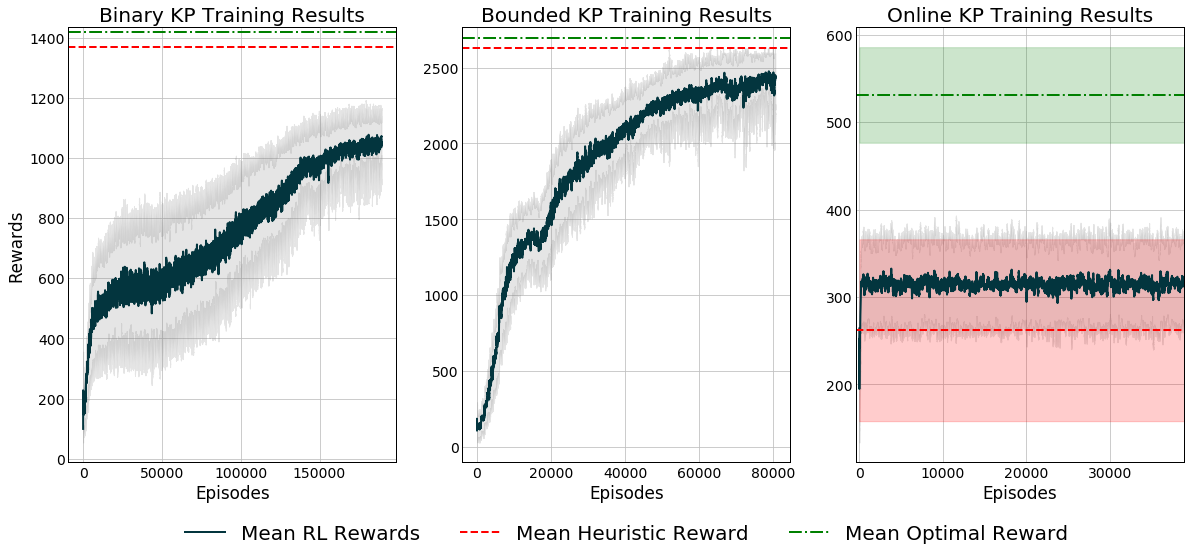}
    \caption{Training curves for three variations of the Knapsack problem. Shaded areas indicate variance.}
    \label{fig:knapsack_training_curves}
\end{figure}

Overall, the RL algorithms do perform well and show they can solve the traditional Knapsack Problem and its variations.
For simple, off-line cases it seems best to resort to heuristics or optimal solutions rather than RL.
However, for stochastic, online cases, there may be benefits in pursuing a reinforcement learning solution as shown in Figure \ref{fig:knapsack_training_curves} (see \ref{sec:model_sizes} for model sizes and computation times).

\section{Virtual Machine Packing}

The Bin Packing Problem (BP) is a classic problem in operations research. 
In its most common form, there are a series of items in a list $i \in I$, each with a given size, $s_i$.
The algorithm must then place these items into one of a potentially infinite number of bins with a size $B$ \citep{Coffman2013}. 
The objective minimizes the number of bins used, the unused space in the bins, or some other, related objective, while ensuring that the bins stay within their size constraints.
For problems with identical bin sizes, we can formulate the problem as:
%
%
\begin{subequations}
\begin{gather}
    \min_{\mathbf{x}, \mathbf{y}} \; \sum_{j=1}^n y_j \label{eq:bin_packing_count_obj} \\
    \textrm{s.t.} \sum_{i=1}^m s_i x_{ij}\leq By_j \quad \forall j \in J \\
    \sum_{j=1}^n x_{ij} = 1 \quad \forall i \in I \\
    x_{ij}, y_{j} \in \{0, 1\} 
\end{gather}
\end{subequations}
where $x_{ij}$ are binary assignment variables for each item that must be assigned to a given bin, and where $y_j$ are bins to be opened by the packing algorithm.

Applications of BPs are common in numerous fields, from loading pallets \citep{Ram1992}, to robotics, box packing \citep{Courcoubetis1990}, stock cutting \citep{Gilmore1961}, logistics, and data centers \citep{Song2014}.

BP's are primarily divided into groups based on their dimensionality, although even 1-D problems are NP-hard (\citet{Christensen2017}, \citet{Johnson1974}). 
1-D problems only consider a size or weight metric, whereas multi-dimensional problems consider an item's area, volume, or combination of features. 

\subsection{Problem Formulation}

We provide multiple versions of the bin packing problem in the OR-Gym library, including the environments implemented by \citet{Balaji2019}, which rely on the examples and heuristic algorithms provided in \citet{Gupta2012}. 
We refer readers to their work for details and results.
In addition to this environment, we implement a multi-dimensional version of the bin packing problem as applied to virtual machines with data from \citet{Cortez2017}. 
This data was collected over a 30-day period from Microsoft Azure data centers containing multiple physical machines (PM) to host virtual machine (VM) instances.
Each VM has certain compute and memory requirements, and the algorithm must map a VM instance to a particular PM without exceeding either the compute or memory requirement. 
In this case, we model the normalized demand at 20-minute increments over a 24-hour period, whereby a new VM instance must be assigned every 20 minutes.

The environment requires mapping of each VM instance to one of 50 PMs at each time step. 
The objective is to minimize unused capacity on each PM with respect to both the compute and memory dimensions for each time step. 
The episode runs for a single, 24-hour period and ends if the model exceeds the limitations of a given PM.
At each time step, the agent can select from one of the 50 PMs available.
If a selection causes a PM to become overloaded, the agent incurs a large penalty and the episode ends.
It is important to ensure that the penalty is sufficiently large to prevent the agent from ending the episode prematurely, otherwise the agent will find a locally optimal strategy by simply ending the episode as quickly as possible.

For the RL implementation, we define the state with vectors providing information on the status of each physical machine (on/off), the current compute and memory loads on each machine, and the incoming demand to be packed.
For this model, we provide results for two RL versions as well, one with action masking and one without.
Action masking is used as a way to enforce capacity constraints by restricting actions that would cause the agent to overload a given PM and thus lead to an abrupt end to the episode and a large negative reward.
This has the effect of reducing the search space for the agent.

\subsection{OR Methods}

Heuristics such as \textit{Best Fit} (BF) \citep{Johnson1974}, and \textit{Sum of Squares} (SS) \citep{Csirik2006}, have been proposed and are well studied, providing theoretical optimality bounds in the limit for one-dimensional BPs. 
For multi-dimensional problems, other heuristics such as \textit{Next Fit Decreasing Height} (NFDH) and \textit{First Fit Decreasing Height} (FFDH) are early and common heuristics approximation ratios of 3 and 2.7, respectively \citep{Christensen2017}. 
Currently, \citet{Bansal2016} provides the best results in the 2-D case with an asymptotic approximation guarantee of $\approx 1.405$.
We employ the \textit{First Fit} algorithm for our VM packing case, which was shown to have an asymptotic performance ratio of 1.7 \citep{Baker1983}.

\begin{algorithm}[!ht]
\begin{algorithmic}[1]
\Require Initialize environment with $J$ empty bins.
\For{Each item $i \in I$ in order}: 
    \For{Each open bin $j \in J$ in order}: 
        \If{$i$ fits in bin $j$}: 
            \State{Place item $i$ in bin $j$ }
        \EndIf 
    \EndFor 
    \If{$i$ does not fit into any open bin}: 
        \State{Open new bin. Place item into new bin.}
    \EndIf 
\EndFor
\end{algorithmic}
\caption{\footnotesize FirstFit VM Packing Algorithm}
\label{algo:ff_vmpacking}
\end{algorithm}

The optimization model solves the VM on shrinking time horizon where the first step optimizes for all time $T$, the second for all time from $t_1$ to $T$ and so forth, until the end of the simulation. 
The decisions in the previous time step are then fixed as it moves forward through time and new demand is being made available.
As in the standard case, the objective is to assign jobs as compactly as possible across the data center.
To make this consistent with the RL formulation, we write this as maximizing a negative value, namely the sum of the occupied space across both dimensions of interest on each physical machine in use, minus the total capacity for each active physical machine.
The full model is given below.
\begin{subequations}
\begin{gather}
    \max_{\mathbf{x}, \mathbf{z}} \; \sum_{t=1}^T  \sum_{p=1}^P \bigg( \sum_{v=1}^V x_{pvt} (D_vj^{mem} + D_v^{cpu} \big) - 2 z_{pt} \bigg)
    \label{eq:vm_packing_obj} \\
    \textrm{s.t.} \sum_{v=1}^V D_v^{mem} x_{pvt} \leq B^{mem} \quad \forall p \in P, t \in T \\
    \sum_{v=1}^V D_v^{cpu} x_{pvt}  \leq B^{cpu} \quad \forall p \in P, t \in T \\
    \sum_{t=1}^T x_{pvt} \leq \tau y_{pv} \quad \forall p \in P, v \in V \\
    \sum_{p=1}^P y_{pv} = 1 \quad \forall v \in V \\
    z_{pt} \geq x_{pvt} \quad \forall p \in P, v \in V, t \in T \\
    x_{pvt}, y_{pv}, z_{pt} \in {0, 1}
\end{gather}
\end{subequations}
where $p$, $v$, and $t$ are indices that denote the physical machine in the set $P$, the virtual machine requirements in $V$, and the time step in $T$ respectively. $D_v^{mem}$ and $D_v^{cpu}$ indicate the memory and CPU demand for each VM instance, which lasts for time $\tau$. Each physical machine may host multiple processes simultaneously, but these are restricted by normalized memory and compute capacities, denoted by $B^{mem}$ and $B^{cpu}$ respectively. 

The model also makes use of multiple binary, decision variables, $x_{pvt}$, $y_{pv}$, and $z_{pt}$. $x_{pvt}$ maps VMs to PMs for each time period. $y_{pv}$ relates VMs to PMs, while $z_{pt}$ enables the time periods of the PMs to be linked and used in the objective.

\subsection{Results}

The RL agent with masking quickly reaches peak performance with rewards hovering around -500 per episode.
This surpasses the results for the heuristic benchmark by about 9\%, but underperforms the more computationally intensive shrinking horizon MILP by 16\%.
These three methods show very similar variance.

\begin{table}[!h]
\centering
\label{tab:vm_results}
\caption{Comparison of RL versus heuristic model and shrinking horizon model solutions. All results are averaged over 100 episodes.}
\ra{1.3}
\scalebox{0.75}{
\begin{tabular}{@{}lcccc@{}}\toprule
VM Packing & RL (No Masking) & RL (Masking) & Heuristic & MILP \\
\midrule
Mean Rewards       & -1,040 & -511 & -556 & -439 \\
Standard Deviation &     17 & 110 &  111 &  111 \\
Performance Ratio vs MILP & 2.37 &  1.16 & 1.27 & 1 \\
\bottomrule
\end{tabular}
}
\end{table}

\begin{figure}[ht]
    \centering
    \includegraphics[width=0.5\textwidth]{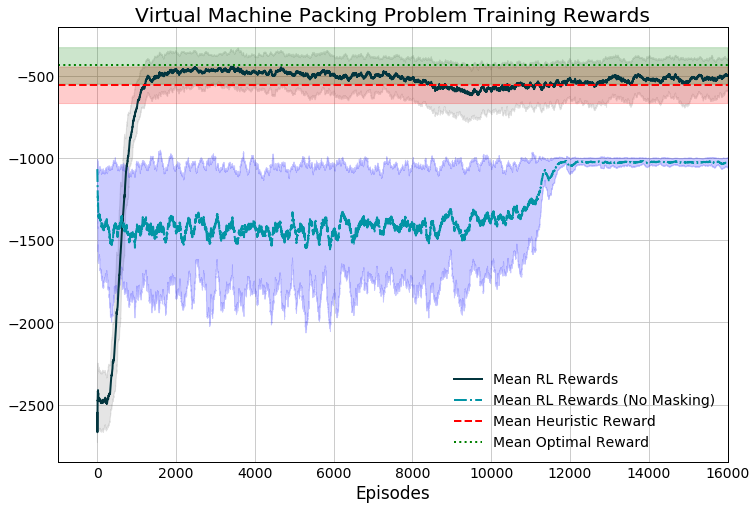}
    \caption{VM Packing Problem with PPO with and without action masking in comparison with the FirstFit heuristic and a shrinking horizon MILP each averaged over 100 episodes.}
    \label{fig:vm_packing}
\end{figure}

As shown in Figure \ref{fig:vm_packing}, the RL agent without masking greatly underperforms relative to all other methods.
The agent was given the same hyperparameters and set identically to the agent with masking, the only difference being that it was able to violate the packing constraints, thereby receiving a large negative reward and immediately ending to the episode.
This example shows the use of a ``soft constraint'' versus a hard constraint in RL set up and design.
The agent without masking quickly learned to take the -1000 reward as fast as possible, thus getting stuck in a local optimum policy because exploration yielded eventual termination by constraint violation and larger negative rewards.
This behavior could be mitigated with further hyperparameter tuning, environment design, or other techniques, but it seems far more effective to simply mask actions that would violate constraints and let the agent learn from there.

\section{Supply Chain Inventory Management}

Managing inventory levels is critical to supply chain sustainability. 
A clear relation exists between inventory levels and order fulfillment service levels (i.e. the more inventory you have, the better you can satisfy your customer's requests). 
However, high inventory levels come at a cost, referred to as holding costs. 
The key is to strike a balance between the trade-off between service level and holding costs. 
In this section, we provide two variations of a Multi-Echelon Inventory Management problem for optimization and training RL agents.

In these inventory management problems (IMPs), a retailer faces uncertain consumer demand from day to day, and must hold inventory at a cost in order to meet that demand.
If the retailer fails to meet that demand, it will either be marked as a backlog order, whereby it may be fulfilled at a later date and lower profit (InvManagement-v0), or simply chalked up as a lost sale with zero profit (InvManagement-v1).
Each day, the retailer must decide how much inventory to purchase from its distributor, who will manufacture and ship the product to the retailer with a given lead time. 
In the multi-echelon case, the distributor will have a supplier, which may also have another supplier above it, and so forth until the supply chain terminates at the original party that consumes the raw materials required for the product, $M$ steps away from the retailer.

In a decentralized supply chain, coordination is key to effectively managing the supply chain. Each stage faces uncertainty in the amount of material requested by the stage succeeding it. A lack of inter-stage coordination can result in bullwhip effects \citep{Lee1997InformationEffect}. In the IMP, each stage operates according to its own unique costs, constraints, and lead times.
The challenge is then to develop a re-order policy for each of the participants in the supply chain to minimize costs and maintain steady operations.

\begin{figure}[ht]
    \centering
    \includegraphics[width=0.45\textwidth]{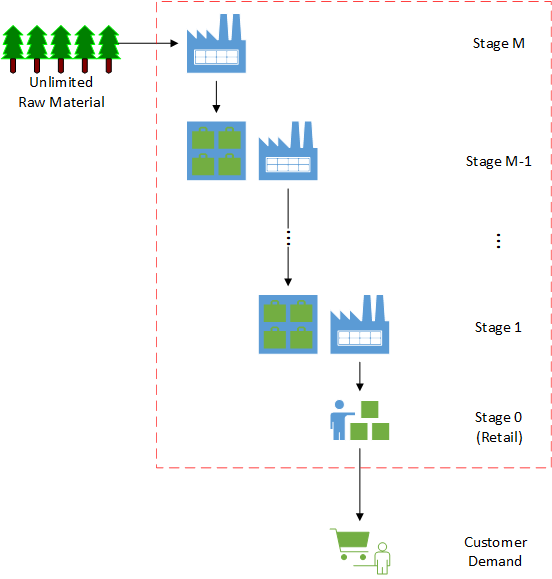}
    \caption{Multi-echelon supply chain.}
    \label{fig:supplychain}
\end{figure}

The IMP environments presented here are based on the work by \citet{Glasserman1995SensitivitySystems}.
In this work, a multi-echelon system with both inventory holding areas and capacitated production areas for each stage is used. 
The inventory holding areas store intermediates that are transformed into other intermediates or final products in the respective production areas.
The default configuration is to have both inventory and production areas at each stage in the supply chain, except for the retailer, which only holds final product inventory, and the supplier furthest upstream, which has virtually unlimited access to the raw material.
However, production areas can be removed if desired by setting the production cost to a large value at those stages.

The standard approach taken in industry to improve the performance in such systems is the use of IPA (infinitesimal perturbation analysis) to determine the optimal parameters for the desired inventory policy.
Although it is acknowledged that the base stock policy may not be optimal for multi-stage capacitated systems, its simplicity makes it attractive for practical implementations. 

Relevant literature addressing other approaches to the IMP are those of \citet{Bertsimas2006ATheory}, \citet{Chu2015Simulation-basedUncertainty}, and \citet{Mortazavi2015}. 
In the work by \citet{Bertsimas2006ATheory}, a general optimization methodology is proposed using robust optimization techniques for both capacitated and uncapacitated systems. 
Their work includes capacity limits on the orders and inventory, but not on production. 
Their model shows benefits in terms of tractability and is solved as either a linear program (LP) or a mixed-integer program (MIP), depending on whether fixed costs are included or not.
\citet{Chu2015Simulation-basedUncertainty} use agent based modeling coupled with a cutting plane algorithm to optimize a multi-echelon supply chain with an ($r$, $Q$) reorder policy under a simulated environment. 
Monte Carlo simulations are used to determine expectations followed by hypothesis testing to deal with the effects of noise when accepting the improvements.
\citet{Mortazavi2015} develop a four-echelon supply chain model with a retailer, distributor and manufacturer. 
They apply Q-learning to learn a dynamic policy to re-order stock over a 12-week cycle with non-stationary demand drawn from a Poisson distribution. 
Several papers in the area of inventory optimization are also available. Of note are those by \citet{Eruguz2016AOptimization} and \citet{Simchi-Levi2012PerformanceSurvey}.

Variations of the IMP include single period and multi-period systems, as well as single product and multi-product systems. The IMP environments currently available (InvManagement-v0 and InvManagement-v1) support single product systems with with stationary demand and either single or multiple time periods. It is assumed that the product is non-perishable and sold in discrete quantities. A depiction of the multi-echelon system is given in Figure \ref{fig:supplychain}.
Stage 0 is the retail site, which is an inventory location that sells the final product to external customers. 
As mentioned previously, stages 1 through $M-1$ have both an inventory area and a manufacturing area, and stage $M$ has only a manufacturing area.
For each unit of inventory transformed, one unit of intermediate or final product is obtained. 
Material produced at a stage is shipped to the inventory area of the stage succeeding it.
Lead times may exist in the production/transfer of material between stages. 
Each manufacturing site has a limited production capacity. 
Each inventory holding area also has a limited holding capacity. 
It is assumed that the last stage has immediate access to an unlimited supply of raw materials and thus a bounded inventory area is not designated for this stage.

\subsection{Problem Formulation}

At each time period in the IMP, the following sequence of events occurs:
\begin{enumerate}
    \item Stages 0 through $M-1$ place replenishment orders to their respective suppliers. Replenishment orders are filled
        according to available production capacity and available inventory at the respective suppliers. Lead times between stages include both production times and transportation times.
    \item Stages 0 through $M-1$ receive incoming inventory replenishment shipments that have made it down the product pipeline
        after the associated lead times have passed.
    \item Customer demand occurs at stage 0 (the retailer) and is filled according to the available inventory at that stage. 
    \item One of the following occurs at each stage,
        \begin{enumerate}
            \item Unfulfilled sales and replenishment orders are backlogged at a penalty. 
            Note: Backlogged sales take priority in the following period.
            \item Unfulfilled sales and replenishment orders are lost with a goodwill loss penalty.
        \end{enumerate}
    \item Surplus inventory is held at each stage at a holding cost.
    \item Any inventory remaining at the end of the last period is lost.
\end{enumerate}
\begin{subequations}
\begin{align}
    & I^m_{t+1} = I^m_t + R^m_{t-L_m} - S^m_t && \forall m \in \mathcal{M}, t \in \mathcal{T} \label{eq:invbal} \\
    & T^m_{t+1} = T^m_t - R^m_{t-L_m} + R^m_t && \forall m \in \mathcal{M}, t \in \mathcal{T} \label{eq:pipebal} \\
    & R^m_t = \min\left(c^{m+1},I^{m+1}_t,\hat{R}^m_t\right) && \forall m \in \mathcal{M} \setminus \{|\mathcal{M}|\}, t \in \mathcal{T} \label{eq:reorder} \\
    & S^m_t = 
    \begin{cases}
        R^{m-1}_t & \text{if $m>0$}\\
        \min\left(I^0_t + R^0_{t-L_m},D_t + B^0_{t-1}\right) & \text{if $m=0$}
    \end{cases}
    && \forall m \in \mathcal{M}, t \in \mathcal{T} \label{eq:sales} \\
    & U^m_t = \hat{R}^{m-1}_t - S^m_t && \forall m \in \mathcal{M}, t \in \mathcal{T} \label{eq:unfulfilled} \\
    & P^m_t = (\alpha)^n \cdot \left(p^m S^m_t -  r^m R^m_t - k^m U^m_t - h^m I^m_{t+1}\right) && \forall m \in \mathcal{M}, t \in \mathcal{T} \label{eq:profit}
\end{align}
\end{subequations}
Six equations govern the behavior of the IMP. 
The first two (Equations \ref{eq:invbal} - \ref{eq:pipebal}) are material balances for the on hand ($I$) and pipeline ($T$) inventory at the beginning of each period $n$, respectively. 
The on-hand inventory at each stage $m$ in the beginning of the next period is equal to the initial inventory in the current period, plus the reorder quantity arrived ($R$, placed $t-L_m$ periods ago), minus the sales in the current period ($S$). 
The pipeline inventory at each stage in the beginning of the next period is equal to the pipeline inventory in the current period, minus the delivered reorder quantity (placed $t-L_m$ periods ago), plus the current reorder quantity placed. 
Equation \ref{eq:reorder} relates the accepted reorder quantity ($R$) to the requested reorder quantity ($\hat{R}$). 
In the absence of production capacity ($c$) and inventory constraints in the stage above the stage $m$, the accepted reorder quantity would be equal to the requested order quantity. 
However, when these constraints exist, they place an upper bound on the reorder quantity that is accepted. 
At stage $M-1$, the term $I^{m+1}_t$ can be ignored or set to $\infty$ since it is assumed that the inventory of raw materials at stage $M$ is infinite.

Equation \ref{eq:sales} gives the sales $S$ at each period, which equal the accepted reorder quantities from the succeeding stages for stages $1$ through $M$ and equal the fulfilled customer demands for the retailer (stage $0$). 
The fulfilled customer demand is equal to the demand plus the previous period's backlog, unless there is insufficient on hand inventory at the retailer, in which case, all of the inventory on hand is sold. 
The unfulfilled demand $U$ or unfulfilled reorder requests are given by Equation \ref{eq:unfulfilled}. 
It should be noted that $\hat{R}^{-1}_t \equiv D_t + B_{t-1}^0$. 
The profit $P$ is given by Equation \ref{eq:profit}, which discounts the profit (sales revenue minus procurement costs, unfulfilled demand costs, and excess inventory holding costs) with a discount factor $\alpha$. $p$, $r$, $k$, and $h$ are the unit sales price, unit procurement cost, unit penalty for unfulfilled demand, and unit inventory holding cost at each stage $m$, respectively. 
For stage $M$, $R^m_t \equiv S^m_t$ and $r^m$ represents a raw material procurement cost. 
If backlogging is not allowed, any unfulfilled demand or procurement orders are lost sales and all $B$ terms are set to 0.

The supply chain inventory management problems were modeled as MDPs, whereby the agent must decide how much stock to re-order from the higher levels at each time step. At each time step $t$, an action $a_{t}^m \in A$ is taken at each stage $m$. The action corresponds to each reorder quantity at each stage in the supply chain. The actions are integer values and maintain the supply capacity and inventory constraints of the form $A \leq C$.

\begin{figure}
    \centering
    \includegraphics[width=0.7\textwidth]{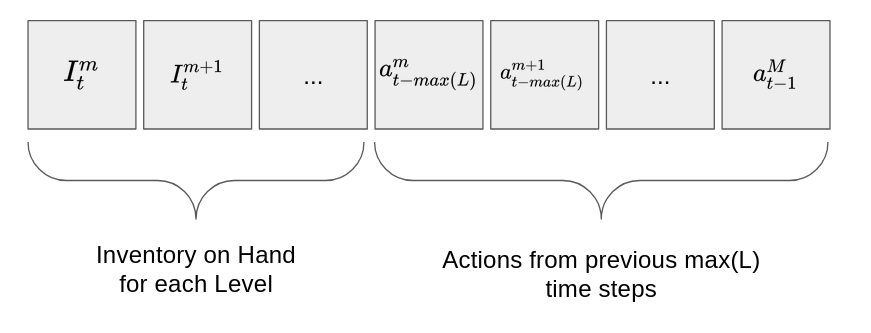}
    \caption{Illustration of state for inventory management environments where $\textrm{max}(L)$ denotes the system's maximum lead time.}
    \label{fig:inv_management_state}
\end{figure}

The states are denoted by the inventory on hand for each level, as well as the previous actions for each of the $\textrm{max}(L)$ time steps in order to capture the inventory in the pipeline (see Figure \ref{fig:inv_management_state}).

The RL agent seeks to maximize the profit of the supply chain as defined in the reward function given in Equation \ref{eq:profit}. The simulation lasts for 30 periods (days), and transitions from one state to the next as material transfers are performed through the supply chain and orders are fulfilled at the retailer.

\subsection{OR Methods}

The base stock policy has been shown to be optimal for capacitated production-inventory systems under certain conditions \citep{Kapuscinski1999OptimalSystems}. 
For multi-stage systems, these conditions are that backlogging is allowed (no lost sales), lead times are fixed, and the capacity at a stage does not exceed the capacity at the stage below it.
Although the base-stock policy is not necessarily optimal under other conditions, it one of the valid OR approaches used in practice due to its simplicity. 
Under this policy, the requested reorder quantity is given by Equation \ref{eq:basestock}, where $z^m$ is the base-stock level at stage $m$ and the term in the summation is the current inventory position at the beginning of period $t$.  
\begin{equation}
    \hat{R}^m_t = \max\left(0,z^m - \sum_{m^\prime = 1}^m \left(I^{m^\prime}_t + T^{m^\prime}_t - B^{m^\prime}_{t-1}\right)\right) \qquad \forall m \in \mathcal{M}, t \in \mathcal{T}
    \label{eq:basestock}
\end{equation}
\cite{Glasserman1995SensitivitySystems} propose a numerical method to determine the optimal base-stock level called infinitesimal perturbation analysis (IPA). 
IPA is a gradient descent approach that minimizes the expected cost over a sample path. 
IPA relies on perturbing a simulated sample path iteratively until the desired improvement in an objective function is obtained. 
Derivatives are calculated or estimated explicitly to provide the updates necessary for the gradient descent. 
To guarantee convergence, the optimization is performed offline. In the IPA implementation for the base stock policy, a simulated sample path of $T$ periods is run with fixed base stock levels. 
The gradients for the state variables (inventory positions, reorder quantities, etc) are determined from the recursive relations in Equations \ref{eq:invbal} - \ref{eq:profit}. 
These are used to update the base stock levels. The updated levels are applied to the same simulated sample path over $T$ periods (iteration number 2). 
The process is repeated until the base stock levels stabilize or the change in the objective function is below a certain tolerance.

\begin{equation}
    f = \frac{1}{|\mathcal{T}|} \mathbf{E}_D \left[\sum_{m\in \mathcal{M}} P^m_t\right]
    \label{eq:samplepath}
\end{equation}

In the present work, we follow this idea of optimizing over a sample path, but apply more robust approaches for the optimization.
Since the objective function for the IMP (normalized expected profit over the sample path, Equation \ref{eq:samplepath}) is non-smooth and non-differentiable for discrete demand distributions, line search algorithms based on the Wolfe conditions \citep{Nocedal2006NumericalOptimization} often fail to converge when determining the step size for the classical IPA approach. 
Instead of settling on taking full step sizes or using an \textit{ad hoc} approach when determining the step sizes, derivative-free optimization (DFO) based on Powell's method \citep{Powell1964AnDerivatives} can be used. 
Another approach available for IMP is that of mixed-integer programming (MIP), which readily expresses the discontinuities in the model equations using binary variables and can guarantee finding the optimal base stock levels over a sample path. 

The benefit of using Powell's method \citep{Powell1964AnDerivatives} is that it does not rely on gradients and can be applied to non-differentiable systems. Since the demand distributions are discrete, Powell's method is followed by either rounding the base stock levels to the nearest integer or performing a local search to find the nearest optimal integer solution. 
In the latter case, enumeration is used to compare all neighboring integer solutions to the optimum found by Powell's method. 
Using Powell's method provides the advantage of being much faster to solve than the MIP alternative, which is NP-hard.

In the MIP approach, the IMP system is modelled as a mixed-integer linear program (MILP) and a MIP solver is used to find the optimal base-stock levels. The disjunctions arising from the minimum operators in Equations \ref{eq:reorder} - \ref{eq:sales} and the maximum operator in Equation \ref{eq:basestock} are reformulated into algebraic inequalities using Big-M reformulations (Equations \ref{eq:max} - \ref{eq:max3}). Furthermore, to ensure the standard multi-echelon condition $z^m \leq z^{m+1} \; \forall m \in \mathcal{M} \setminus |\mathcal{M}|$, the base stock levels are written in terms of the total inventory level, $x^m \geq 0$, at each stage ($z^m = \sum_{m^\prime = 1}^m x^{m^\prime}$).
\begin{subequations}
\begin{gather}
    x = \max(A_1,...,A_{|I|}) \label{eq:max}\\
    A_i \leq x \leq A_i + M_i(1-y_i) \quad \forall i \in I \label{eq:max1} \\
    x = \min(A_1,...,A_{|I|}) \label{eq:min} \\
    A_i + M_i(1-y_i) \leq x \leq A_i \quad \forall i \in I \label{eq:min1} \\
    \sum_{i \in I} y_i = 1 \label{eq:max2} \\
    y \in \{0,1\}^{|I|} \label{eq:max3}
\end{gather}
\end{subequations}

A third approach to the IMP is to use a dynamic reorder policy that is determined by solving a linear program (LP) at each time period with a shrinking horizon (SHLP). This approach requires prior knowledge of the demand probability distribution and assumes the expected value of the demand for all time periods. After solving the optimization for the current time period in the simulation, the optimal reorder action found for the current period is implemented. All future reorder actions are discarded and the process is repeated for the next time period. The shrinking horizon model has no binary variables as a result the removal of Eq \ref{eq:basestock}. In the absence of targeted base stock levels, the requested reorder quantity $\hat{R}_t^m$ becomes the same as the accepted reorder quantity $R_t^m$. As a result, Eq \ref{eq:reorder} can be replaced with $R_t^m \le c^{m+1}$ and $R_t^m \le I_t^{m+1} $ and the second part of Eq \ref{eq:sales} can be replaced with $S_t^0 \le I_t^0+R_{t-L_m}^0$ and $S_t^0 \le D_t+B_{t-1}^0$.

\begin{table}[!ht]
    \centering
    \caption{Parameters values for both Inventory Management Environments}
    \label{tab:nvp_example1}
    \scalebox{1}{
    \begin{tabular}{lccccc}
    \toprule
    Parameter & Symbol & Stage 0 & Stage 1 & Stage 2 & Stage 3 \\ \midrule
    Initial Inventory       & $I_0$     & 100     & 100     & 200     & -       \\
    Unit Sales Price        & $p$       & \$2.00  & \$1.50  & \$1.00  & \$0.75  \\
    Unit Replenishment Cost & $r$       & \$1.50  & \$1.00  & \$0.75  & \$0.50  \\
    Unit Backlog Cost       & $k$       & \$0.10  & \$0.075 & \$0.05  & \$0.025 \\
    Unit Holding Cost       & $h$       & \$0.15  & \$0.10  & \$0.05  & -       \\
    Production Capacity     & $c$       & -       & 100     & 90      & 80      \\
    Lead Times              & $L$       & 3       & 5       & 10      & -       \\ \bottomrule
    \end{tabular}
    }
\end{table}



\subsection{Results}

Two examples are run to compare the OR (DFO, MIP, and SHLP) and RL approaches in the inventory management problem. It should be noted that the DFO and MIP approaches use static base stock policies, whereas the SHLP and RL approaches use dynamic reorder policies. 
Both approaches are compared to an oracle model, which is a LP model that determines the optimal dynamic reorder policy for that run by using the actual demand values for each time period. InvManagement-v0 is a 30 period, 4 stage supply chain with backlog. 
A Poisson distribution is used for the demand with a mean of 20. A discount factor for the time value of money of 97\% percent is used (3\% discount). 
The second problem, InvManagement-v1, is identical to InvManagement-v0, except it does not have a backlog.
The stage specific parameters used in both environments are given in Table \ref{tab:nvp_example1}.

\begin{figure}[!ht]
    \centering
    \includegraphics[width=0.7\textwidth]{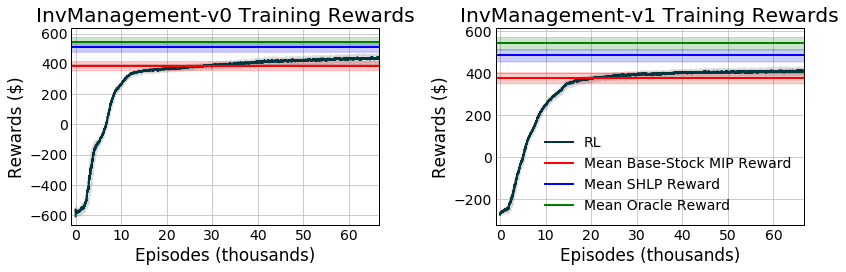}
    \caption{Training curves smoothed over 100 episodes for RL algorithm compared to oracle and MILP base-stock models. Shaded areas indicate standard deviation of rewards.}
    \label{fig:inv_management_training_curves}
\end{figure}

Training curves for both environments, v0 and v1, as well as the mean values for the oracle, SHLP, and MILP models are given in Figure \ref{fig:inv_management_training_curves}.

\begin{figure}
    \centering
    \includegraphics[width=0.5\textwidth]{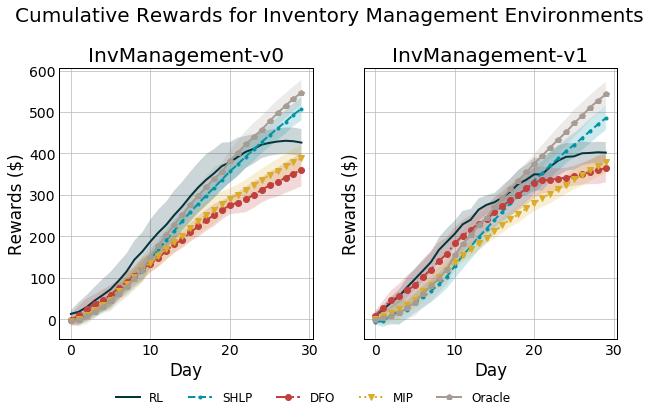}
    \caption{Cumulative profit for both inventory management models. Results averaged over 10 episodes. Shaded areas indicate standard deviation of rewards.}
    \label{fig:inv_management_episode_example}
\end{figure}

\begin{figure}
    \centering
    \includegraphics[width=0.8\textwidth]{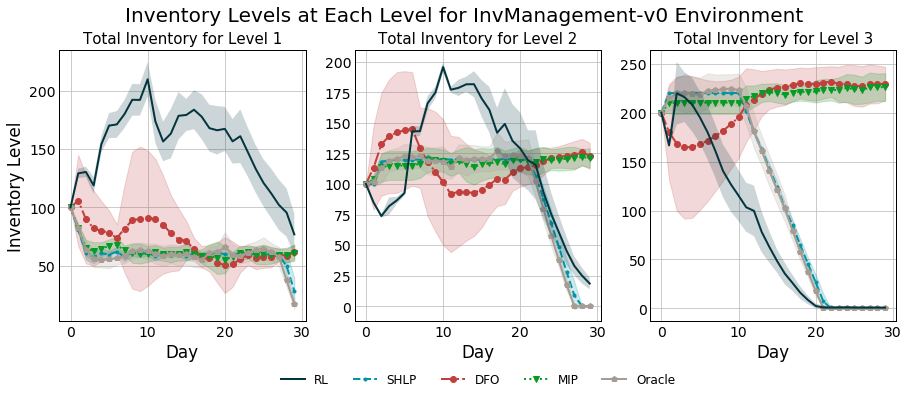}
    \caption{Average inventory on hand at each echelon for InvManagement-v0.}
    \label{fig:invmanagement-v0_avg_echelon_inventory}
\end{figure}

For the InvManagement-v0 environment, the RL model outperformed the static policy models, but was outperformed by the shrinking horizon model, achieving a performance ratio of 1.2 (Table \ref{tab:invmanagement_results}). 
The daily rewards for each of the compared models are given in Figure \ref{fig:inv_management_episode_example}, where it can be seen that the RL model's rewards track closely with the oracle and SHLP models before leveling off towards the end of the episode.
The model does this because it has been trained on a 30-day episode, and, unlike the static base stock policy models, it does not need to continue to order additional inventory towards the end of the run because the holding costs become large and the extra stock will not be sold within the episode horizon to be worthwhile.
This behavior is shown in the inventory plots given in Figure \ref{fig:invmanagement-v0_avg_echelon_inventory}, where the RL inventory levels decrease as the episode progresses, as is also observed with the oracle and SHLP models, which are also aware of the limited horizon.

The static policy models do not exhibit this behavior and continue targeting the optimal base stock levels, which increases the holding costs and causes the profit to suffer. However, in a real application, the supply chain is likely to operate for more than 30 days, and dropping inventories towards the end of the run would be detrimental to the supply chain's performance beyond the 30 days.
The RL model could be trained on a continuous environment that cycles month after month, drawing demand from the same probability distribution to avoid this behavior in a real-world application.

\begin{table}
\centering
\caption{Total reward comparison for InvManagement problems and the various models used to solve them.}
\label{tab:invmanagement_results}
\ra{1.3}
\scalebox{0.75}{
\begin{tabular}{@{}lccccc@{}}\toprule
InvManagement-v0 & RL & SHLP & DFO & MIP & Oracle \\
\midrule
Mean Rewards       & 438.8 & 508.0 & 360.9 & 388.0 & 546.8 \\
Standard Deviation & 30.6 & 28.1 & 39.9 & 30.8  & 30.3 \\
Performance Ratio vs Oracle & 1.2 & 1.1 & 1.5 & 1.4  & 1.0\\
\bottomrule
InvManagement-v1 & RL & SHLP & DFO & MIP & Oracle \\
\midrule
Mean Rewards       & 409.8 & 485.4 & 364.3 & 378.5  & 542.7 \\
Standard Deviation & 17.9 & 29.1 & 33.8 & 26.1 & 29.9 \\
Performance Ratio vs Oracle & 1.3 & 1.1 & 1.5 & 1.4 & 1.0\\
\end{tabular}
}
\end{table}

\begin{figure}
    \centering
    \includegraphics[width=0.8\textwidth]{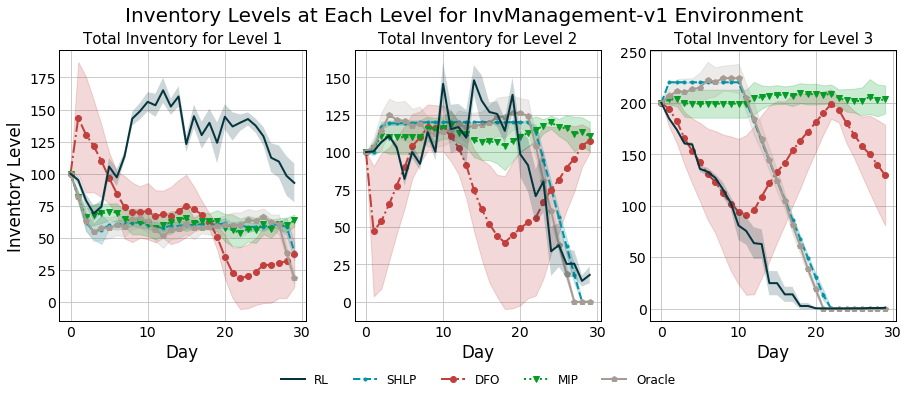}
    \caption{Average inventory on hand at each echelon for InvManagment-v1.}
    \label{fig:invmanagement-v1_avg_echelon_inventory}
\end{figure}

The RL model in InvManagement-v1 (no backlog orders), performs slightly worse relative to the oracle with a performance ratio of 1.3 (Table \ref{tab:invmanagement_results}).
The same end-of-episode dynamics observed in InvManagement-v0 are also visible in Figure \ref{fig:invmanagement-v1_avg_echelon_inventory}.

For both environment variations, the RL model outperforms the static policy models, scoring performance ratios between 1.2-1.3.
This is shows the potential for RL to learn competitive, dynamic policies for multi-stage inventory management problems. 

\section{Asset Allocation}

Asset allocation refers to the task of creating a collection (portfolio) of financial assets as to maximize an investors return, subject to some investment criteria and constraints. The most famous approach to portfolio optimization stems from Markowitz's mean-variance framework from the 1950s that frames portfolio optimization as a trade-off between the return of an asset (modeled using its expected value) and the risk associated with the asset (modeled by its variance) \citep{Markowitz1952}. 
Since then, the asset allocation problem has been extensively explored in the literature with a multitude of models that take into account many different practical considerations (see \citet{Black1992}, \citet{Perold1984}, \citet{Fabozzi2007}, \citet{Konno1991}, \citet{Krokhmal2003}, and \citet{DeMiguel2009}).

\subsection{Problem Formulation}

Here, we focus on the \textit{multi-period asset allocation problem} (MPAA) of \citet{Dantzig1993}. An investor starts with a portfolio $\mathbf{x}^0 = [x^0_1,...,x^0_n]$ consisting of $n$ assets, with an additional quantity of cash we call a ``zero'' asset, $x_0^0$. The investor wants to determine the optimal distribution of assets over the next $L$ periods to maximize total wealth at the end of the final investment horizon of total length $L$.

The optimization variables are $b_i^l$ (the amount of asset $i$ bought at the beginning of period $l$) and $s_i^l$ (the amount sold) where $i=1,...,n$ and $l=1,...,L$. 

The price of an asset $i$ in period $l$ is denoted by $P_i^l$. We assume that the cash account is interest-free meaning that $P_0^l = 1, \forall l$. The sales and purchases of an asset $i$ in period $l$ are given by proportional transaction costs $\alpha_i^l$ and $\beta_i^l$, respectively. These prices and costs are subject to change asset-to-asset and period-to-period. 

The deterministic version of this problem is given by the following formulation:
\begin{subequations}\label{eq:DetPOFormulation}
\centering
\begin{gather}
    \max_{\mathbf{x},\mathbf{s},\mathbf{b}} \sum_{i=0}^n P_i^L x_i^L \label{eq:POobj} \\
    x_0^l \leq x_0^{l-1} + \sum_{i=1}^n (1-\alpha_i^l) P_i^l s_i^l - \sum_{i=1}^n (1+\beta_i^l) P_i^l b_i^l \qquad \forall l \in L \label{eq:PObal} \\
    x_i^l = x_i^{l-1} - s_i^l + b_i^l \qquad \forall i \in n \quad \forall l \in L \label{eq:POsum}\\
    s_i^l, b_i^l, x_i^l \geq 0 \qquad \forall i \in n \quad \forall l \in L\label{eq:POnonneg}
\end{gather}
\end{subequations}
This is a linear programming problem that can be easily solved by common optimization solvers. 

In the RL environment, we consider the optimization of a portfolio derived from Equations \ref{eq:DetPOFormulation}. This portfolio initially contains \$100 in cash and three other assets and our goal is to optimize portfolio value over a time horizon of 10 periods.  
Of course, the prices of each asset for each period are not known in advance. Consequently, these prices are modeled as Gaussian random variables with a fixed mean and variance for each asset in each period. In each instantiation of the environment, these prices take on a new value. Transaction costs are fixed in all instances. 

Each step in the RL environment corresponds to a single investment period. The state of the environment, $s$, at a particular period is described by the following vector of the current period, $l$, as well as cash and asset quantities and prices, corresponding to the notation of (Equations \ref{eq:DetPOFormulation}):
\begin{equation}
s = \big[l, x_0^l, x_1^l, x_2^l, x_3^l, P_1^l, P_2^l, P_3^l\big]
\end{equation}
At each step, the RL agent must decide how much, if any, of each particular asset to buy or sell. The action, $a$, is described by the following vector of amounts of the three assets to buy or sell, corresponding to the notation of from Equations \ref{eq:DetPOFormulation}:
\begin{equation}
a = \big[\delta_1^l, \delta_2^l, \delta_3^l\big]
\end{equation}
where, 
  \[
    \delta_i^l = \left\{\begin{array}{lr}
        s_i^l \quad (b_i^l = 0), & \text{for } \delta_i^l < 0\\
        b_i^l \quad (b_i^l = 0), & \text{for } \delta_i^l > 0\\
        s_i^l, b_i^l=0  & \text{for } \delta_i^l = 0
        \end{array}\right\} 
  \]
  
  $$ \delta_i^l \in \big[-2000, 2000\big]$$
The reward $r$ is given only at the very end of the episode and is equal to the value of the objective function of Equation \ref{eq:POobj} (i.e. the portfolio value). 
An alternative design choice that we considered was to give a reward at each step (for example, the current portfolio value). This implementation did prove to facilitate learning but we chose the sparse-reward formulation to demonstrate the power of RL even in relatively challenging situations. 

\subsection{OR Methods}

Robust optimization (RO) is an established field in operations-research that rigorously addresses decision-making under uncertainty. 
In the robust optimization paradigm, optimization problem parameters are treated as unknown quantities bounded by an uncertainty set.
It is the objective of RO to find good solutions that are feasible no matter what values within the uncertainty set the problem parameters take \citep{Ben-Tal2009}.
In essence, robust optimization allows the practitioner to optimize the worst-case scenario. 

To benchmark the performance of the RL agent, we formulate and solve the corresponding robust-optimization problem as given in Equations \ref{eq:DetPOFormulation}, originally from \cite{Ben-Tal2000}.
Constraints on cash flow are modeled using a $3-\sigma$ approach such that the chosen investment policy will be feasible in 99.7\% of possible scenarios.
The goal of the optimization is to find an investment policy such that the portfolio value will be greater than or equal to the objective function value of the optimization problem.
We refer the reader to \cite{Cornuejols2006} for a clear description and explanation of the robust formulation.  

We then take the investment policy (i.e. the amount of each asset to buy and sell in each period) and simulate this policy for several different realizations of the parameters within the uncertainty set. We use the average rewards achieved as a benchmark against the policy found by the RL agent. 

An alternative to the robust optimization approach would have been to use multi-stage stochastic programming (MSP). To paraphrase the arguments of \cite{Ben-Tal2000}, RO was highlighted instead because: it is much more tractable than the MSP, any simplifying implementation of MSP (e.g. by solving on a rolling-horizon) would be \textit{ad hoc} and would not necessarily yield a superior solution than the RO approach, and, although the RO approach seems `conservative', the mathematical form of the RO is such that the RO solution is not as conservative as it may seem. As important, a central goal of asset allocation is not only to maximize expected rewards, but also to minimize risk. Robust optimization, which minimizes the downside-risk, is a natural approach to achieve portfolios with high returns under a variety of market conditions. As the results of \cite{Ben-Tal2000} show for this problem, not only does RO match the performance of MSP in terms of mean portfolio value, the standard-deviation (and hence the risk) of the return is substantially lower. 

We solve the RO model using BARON 19.7.13 on a 2.9 GHz Intel i7-7820HQ CPU \citep{Sahinidis1996}.

\subsection{Results}

The solution to the deterministic optimization problem, where all prices (equal to their mean values in the RO and RL formulations) are known with certainty, is \$17,956.20. 
However, this value is not practically important because market prices are never known in advance.

The robust optimization solution is a more practically important metric for comparison.
While the uncertainty set chosen is conservative, the RO approach multiplies the initial \$100 in cash to a portfolio value of at least \$610.17 for 99.7\% of the parameter space. 
This value corresponds to the objective function value of the RO problem, and is the worst-case return.  

\begin{figure}[!ht]
    \centering
    \includegraphics[width=0.55\textwidth]{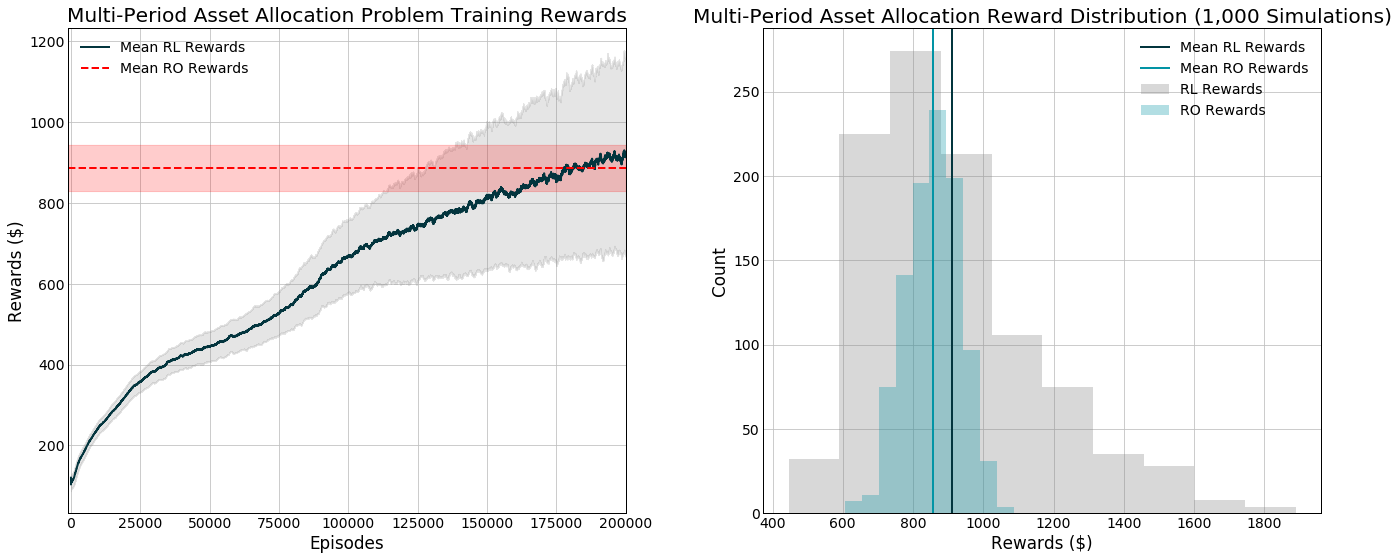}
    \caption{Training rewards for RL algorithm smoothed over 100 episodes and average rewards given by implementing robust optimization solution. Shaded areas indicate standard deviation of rewards. Histogram compares results from 1,000 simulated episodes for RL and RO. Notice the wider distribution and skew of the RL solution which offers less downside protection.}
    \label{fig:mpaa_results}
\end{figure}

The solution to the RO problem corresponds to an investment strategy. 
When we take the RO strategy and actually apply it in $10^3$ randomly generated instances of this problem, we see that the RO policy is able to multiply the initial cash to an average portfolio value of about \$865 with a standard deviation of roughly \$80. 
Results for the RL and RO approaches are given in Figure \ref{fig:mpaa_results}.

We see that, despite the sparse-reward formulation, the RL agent is able to successfully learn a policy that, on average, yields higher portfolio values than the strategy given by RO.
Purely in the sense of maximizing expected portfolio value, RL is a strong alternative to robust optimization.
On the downside however, the variance of the RL portfolio values is higher than in the case of RO.
Consequently, RO seems to be the better alternative when our primary concern is risk, achieving worst-case returns that are much higher than those from RL.
As a final note, the number of episodes required to train the RL agent was very high, resulting in a training time on the order of hours.
On the other hand, the robust optimization policy was found in just a few minutes.

\section{Conclusion}

We have developed a library of reinforcement learning environments consisting of classic operations research and optimization problems. 
RL traditionally relies on environments being formulated as Markov Decision Processes (MPDs).
These are sequential decision processes where a decision or action in one state influences the transition to the subsequent state.
In many optimization problems, sequential decisions are required making RL straightforward to apply to the problem.
Hard constraints can be imposed on RL agents via action masking, whereby the probability associated with selecting an action that would lead to a constraint violation are set to 0 ensuring these actions are not selected.
The use of action masking reduces the search space and can both improve the learned policy and reduce the amount of training time required by the model.

Additionally, we have provided benchmarks for four problem classes showing results for RL, heuristics, and optimization models.
In all cases, RL is capable of learning a competent policy. 
RL outperforms the benchmark in many of the more complex and difficult environments where uncertainty plays a significant role, and may be able to provide value in similar, industrial applications.
RL did not outperform the heuristic models in the off-line knapsack problems.
This does not come as a complete surprise given that the knapsack that the knapsack problem has been well-studied for many decades, and numerous heuristics have been developed to solve the problem and increase the state-of-the-art solutions.
In cases such as this, the additional complexity and computational costs of RL may not be worthwhile, however note that in formulation of the off-line knapsack problems, uncertainty also played no role.

RL provides a general framework to solve these various problems. 
Each system was solved with the same algorithm using three layers and 128 nodes each and minor hyperparameter tuning for the learning rate and entropy loss within a standard library.
With new tools such as Ray, RL becomes increasingly accessible for solving classic optimization problems under uncertainty as well as industrial equivalents where optimization models do not exist, or are too expensive to develop and compute online.


It may be possible to extract rules from the RL policies to develop better heuristic solutions for each of the problems.
Additionally, given the structure of these problems and the fact that optimization models exist for them, it is rife for exploration of hybrid RL and mathematical programming approaches which can combine the speed and flexibility of RL with the solutions available from mathematical programming to improve results and time to solution.
Another interesting avenue to pursue is the use of graph neural networks and RL, which may be able to benefit from the structure of these models.
Finally, the OR-Gym library contains additional environments from the literature for experimentation.\footnote{The full list of envirionments, as well as download and installation instructions can be found here: \url{https://github.com/hubbs5/or-gym}}

\newpage

\bibliography{references.bib} 

\begin{thebibliography}{57}
\providecommand{\natexlab}[1]{#1}
\providecommand{\url}[1]{\texttt{#1}}
\expandafter\ifx\csname urlstyle\endcsname\relax
  \providecommand{\doi}[1]{doi: #1}\else
  \providecommand{\doi}{doi: \begingroup \urlstyle{rm}\Url}\fi

\bibitem[Baker and Schwarz(1983)]{Baker1983}
B.~S. Baker and J.~S. Schwarz.
\newblock {Shelf Algorithms for Two-Dimensional Packing Problems}.
\newblock \emph{SIAM Journal of Computing}, 12\penalty0 (3):\penalty0 508--526,
  1983.

\bibitem[Balaji et~al.(2019)Balaji, Bell-Masterson, Bilgin, Damianou, Garcia,
  Jain, Luo, Maggiar, Narayanaswamy, and Ye]{Balaji2019}
B.~Balaji, J.~Bell-Masterson, E.~Bilgin, A.~Damianou, P.~M. Garcia, A.~Jain,
  R.~Luo, A.~Maggiar, B.~Narayanaswamy, and C.~Ye.
\newblock {ORL: Reinforcement Learning Benchmarks for Online Stochastic
  Optimization Problems}.
\newblock 2019.
\newblock URL \url{http://arxiv.org/abs/1911.10641}.

\bibitem[Bansal et~al.(2016)Bansal, Oosterwijk, Vredeveld, and van~der
  Zwaan]{Bansal2016}
N.~Bansal, T.~Oosterwijk, T.~Vredeveld, and R.~van~der Zwaan.
\newblock {Approximating Vector Scheduling: Almost Matching Upper and Lower
  Bounds}.
\newblock \emph{Algorithmica}, 76\penalty0 (4):\penalty0 1077--1096, 2016.
\newblock ISSN 14320541.
\newblock \doi{10.1007/s00453-016-0116-0}.

\bibitem[Bellman(1957)]{Bellman1957}
R.~Bellman.
\newblock {A Markovian decision process}, 1957.
\newblock ISSN 01650114.

\bibitem[Ben-Tal et~al.(2000)Ben-Tal, Margalit, and Nemirovski]{Ben-Tal2000}
A.~Ben-Tal, T.~Margalit, and A.~Nemirovski.
\newblock {Robust modeling of multi-stage portfolio problems}.
\newblock In \emph{High Performance Optimization}, pages 303--323. 2000.

\bibitem[Ben-Tal et~al.(2009)Ben-Tal, El~Ghaoui, and Nemirovsky]{Ben-Tal2009}
A.~Ben-Tal, L.~El~Ghaoui, and A.~Nemirovsky.
\newblock \emph{{Robust Optimization}}.
\newblock Princeton University Press, 2009.

\bibitem[Berner et~al.(2019)Berner, Brockman, Chan, Cheung, Dennison, Farhi,
  Fischer, Hashme, Hesse, J{\'{o}}zefowicz, Gray, Olsson, Pachocki, Petrov,
  Pond{\'{e}}~de Oliveira~Pinto, Raiman, Salimans, Schlatter, Schneider, Sidor,
  Sutskever, Tang, Wolski, and Zhang]{Berner2019a}
C.~Berner, G.~Brockman, B.~Chan, V.~Cheung, C.~Dennison, D.~Farhi, Q.~Fischer,
  S.~Hashme, C.~Hesse, R.~J{\'{o}}zefowicz, S.~Gray, C.~Olsson, J.~Pachocki,
  M.~Petrov, H.~Pond{\'{e}}~de Oliveira~Pinto, J.~Raiman, T.~Salimans,
  J.~Schlatter, J.~Schneider, S.~Sidor, I.~Sutskever, J.~Tang, F.~Wolski, and
  S.~Zhang.
\newblock {Dota 2 with Large Scale Deep Reinforcement Learning}.
\newblock Technical report, 2019.
\newblock URL \url{https://www.facebook.com/OGDota2/}.

\bibitem[Bertsimas and Thiele(2006)]{Bertsimas2006ATheory}
D.~Bertsimas and A.~Thiele.
\newblock {A robust optimization approach to inventory theory}.
\newblock \emph{Operations Research}, 54\penalty0 (1):\penalty0 150--168, 1
  2006.
\newblock ISSN 0030364X.
\newblock \doi{10.1287/opre.1050.0238}.
\newblock URL
  \url{http://pubsonline.informs.org:150-168.https://doi.org/10.1287/opre.1050.0238http://www.informs.org}.

\bibitem[Black and Litterman(1992)]{Black1992}
F.~Black and R.~Litterman.
\newblock {Global portfolio optimization}.
\newblock \emph{Financial Analysts Journal}, 48\penalty0 (5), 1992.

\bibitem[Brockman et~al.(2016)Brockman, Cheung, Pettersson, Schneider,
  Schulman, Tang, and Zaremba]{Brockman2016}
G.~Brockman, V.~Cheung, L.~Pettersson, J.~Schneider, J.~Schulman, J.~Tang, and
  W.~Zaremba.
\newblock {OpenAI Gym}.
\newblock pages 1--4, 2016.
\newblock ISSN 00217298.
\newblock \doi{10.1241/johokanri.44.113}.
\newblock URL \url{http://arxiv.org/abs/1606.01540}.

\bibitem[Christensen et~al.(2017)Christensen, Khan, Pokutta, and
  Tetali]{Christensen2017}
H.~I. Christensen, A.~Khan, S.~Pokutta, and P.~Tetali.
\newblock {Approximation and online algorithms for multidimensional bin
  packing: A survey}.
\newblock \emph{Computer Science Review}, 24:\penalty0 63--79, 2017.
\newblock ISSN 15740137.
\newblock \doi{10.1016/j.cosrev.2016.12.001}.
\newblock URL \url{http://dx.doi.org/10.1016/j.cosrev.2016.12.001}.

\bibitem[Chu et~al.(2015)Chu, You, Wassick, and
  Agarwal]{Chu2015Simulation-basedUncertainty}
Y.~Chu, F.~You, J.~M. Wassick, and A.~Agarwal.
\newblock {Simulation-based optimization framework for multi-echelon inventory
  systems under uncertainty}.
\newblock \emph{Computers and Chemical Engineering}, 73:\penalty0 1--16, 2
  2015.
\newblock ISSN 00981354.
\newblock \doi{10.1016/j.compchemeng.2014.10.008}.

\bibitem[Coffman et~al.(2013)Coffman, Csirik, Galambos, Martello, and
  Vigo]{Coffman2013}
E.~G. Coffman, J.~Csirik, G.~Galambos, S.~Martello, and D.~Vigo.
\newblock {Bin Packing Approximation Algorithms: Survey and Classification}.
\newblock In \emph{Handbook of Combinatorial Optimization}, pages 455--531.
  Springer, Heidelberg, 2013.

\bibitem[Cornuejols and Tutuncu(2006)]{Cornuejols2006}
G.~Cornuejols and R.~Tutuncu.
\newblock {Chapter 20: Robust Optimization Models in Finance}.
\newblock In \emph{Optimization Methods in Finance}, pages 309--312. 2006.

\bibitem[Cortez et~al.(2017)Cortez, Bonde, Muzio, Russinovich, Fontoura, and
  Bianchini]{Cortez2017}
E.~Cortez, A.~Bonde, A.~Muzio, M.~Russinovich, M.~Fontoura, and R.~Bianchini.
\newblock {Resource Central: Understanding and Predicting Workloads for
  Improved Resource Management in Large Cloud Platforms}.
\newblock In \emph{Proceedings of the International Symposium on Operating
  Systems Principles (SOSP)}, 2017.
\newblock ISBN 9781450350853.
\newblock \doi{10.1097/01.nnn.0000365457.75237.2b}.
\newblock URL
  \url{https://www.microsoft.com/en-us/research/publication/resource-central-understanding-predicting-workloads-improved-resource-management-large-cloud-platforms/}.

\bibitem[Courcoubetis and Weber(1990)]{Courcoubetis1990}
C.~Courcoubetis and R.~Weber.
\newblock {Stability of on-line bin packing with random arrivals and
  long-run-average constraints}.
\newblock \emph{Probability in the Engineering and Informational Sciences},
  4\penalty0 (4):\penalty0 447--460, 1990.
\newblock ISSN 14698951.
\newblock \doi{10.1017/S0269964800001753}.

\bibitem[Csirik et~al.(2006)Csirik, Johnson, Kenyon, Orlin, Shor, and
  Weber]{Csirik2006}
J.~Csirik, D.~S. Johnson, C.~Kenyon, J.~B. Orlin, P.~W. Shor, and R.~R. Weber.
\newblock {On the Sum-of-Squares algorithm for bin packing}.
\newblock \emph{Journal of the ACM}, 53\penalty0 (1):\penalty0 1--65, 2006.
\newblock ISSN 00045411.
\newblock \doi{10.1145/1120582.1120583}.

\bibitem[Dantzig(1957)]{Dantzig1957}
G.~B. Dantzig.
\newblock {Discrete-Variable Extremum Problems}.
\newblock \emph{INFORMS}, 5\penalty0 (2):\penalty0 266--277, 1957.

\bibitem[Dantzig and Infanger(1993)]{Dantzig1993}
G.~B. Dantzig and G.~Infanger.
\newblock {Multi-stage stochastic linear programs for portfolio optimization}.
\newblock \emph{Annals of Operations Research}, 45:\penalty0 59--76, 1993.

\bibitem[DeMiguel et~al.(2009)DeMiguel, Garlappi, Nogales, and
  Uppal]{DeMiguel2009}
V.~DeMiguel, L.~Garlappi, F.~J. Nogales, and R.~Uppal.
\newblock {A Generalized Approach to Portfolio Optimization: Improving
  Performance by Constraining Portfolio Norms}.
\newblock \emph{Management Science}, 55\penalty0 (9), 2009.

\bibitem[Eruguz et~al.(2016)Eruguz, Sahin, Jemai, and
  Dallery]{Eruguz2016AOptimization}
A.~S. Eruguz, E.~Sahin, Z.~Jemai, and Y.~Dallery.
\newblock {A comprehensive survey of guaranteed-service models for
  multi-echelon inventory optimization}, 2 2016.
\newblock ISSN 09255273.

\bibitem[Fabozzi et~al.(2007)Fabozzi, Kolm, Pachamanova, and
  Focardi]{Fabozzi2007}
F.~J. Fabozzi, P.~N. Kolm, D.~A. Pachamanova, and S.~M. Focardi.
\newblock \emph{{Robust Portfolio Optimization and Management}}.
\newblock John Wiley {\&} Sons, 2007.

\bibitem[Gilmore and Gomory(1961)]{Gilmore1961}
P.~C. Gilmore and R.~E. Gomory.
\newblock {A Linear Programming Approach to the Cutting-Stock Problem}.
\newblock \emph{Operations Research}, 9\penalty0 (6):\penalty0 849--859, 1961.
\newblock ISSN 0030-364X.
\newblock \doi{10.1287/opre.9.6.849}.

\bibitem[Glasserman and Tayur(1995)]{Glasserman1995SensitivitySystems}
P.~Glasserman and S.~Tayur.
\newblock {Sensitivity analysis for base-stock levels in multiechelon
  production-inventory systems}.
\newblock \emph{Management Science}, 41\penalty0 (2):\penalty0 263--281, 1995.

\bibitem[Gupta and Radovanovic(2012)]{Gupta2012}
V.~Gupta and A.~Radovanovic.
\newblock {Online Stochastic Bin Packing}.
\newblock pages 1--21, 2012.
\newblock URL \url{http://arxiv.org/abs/1211.2687}.

\bibitem[{Gurobi Optimization LLC}(2018)]{Gurobi2018}
{Gurobi Optimization LLC}.
\newblock {Gurobi}, 2018.
\newblock URL \url{https://www.gurobi.com}.

\bibitem[Han et~al.(2015)Han, Kawase, and Makino]{Han2015}
X.~Han, Y.~Kawase, and K.~Makino.
\newblock {Randomized algorithms for online knapsack problems}.
\newblock \emph{Theoretical Computer Science}, 562\penalty0 (C):\penalty0
  395--405, 2015.
\newblock ISSN 03043975.
\newblock \doi{10.1016/j.tcs.2014.10.017}.
\newblock URL \url{http://dx.doi.org/10.1016/j.tcs.2014.10.017}.

\bibitem[Hart et~al.(2017)Hart, Laird, Woodruff, Hackebeil, Nicholson, and
  Siirola]{Hart2017}
W.~E. Hart, C.~D. Laird, D.~L. Woodruff, G.~A. Hackebeil, B.~L. Nicholson, and
  J.~D. Siirola.
\newblock \emph{{Pyomo — Optimization Modeling in Python}}.
\newblock Springer, Cham, Switzerland, 2nd edition, 2017.
\newblock ISBN 9783319588193.

\bibitem[Hubbs et~al.(2020)Hubbs, Li, Sahinidis, Grossmann, and
  Wassick]{Hubbs2020}
C.~D. Hubbs, C.~Li, N.~V. Sahinidis, I.~E. Grossmann, and J.~M. Wassick.
\newblock {A deep reinforcement learning approach for chemical production
  scheduling}.
\newblock \emph{Computers and Chemical Engineering}, 2020.
\newblock \doi{https://doi.org/10.1016/j.compchemeng.2020.106982}.

\bibitem[Johnson(1974)]{Johnson1974}
D.~S. Johnson.
\newblock {Approximation algorithms for combinatorial problems}.
\newblock \emph{Journal of Computer and System Sciences}, 9\penalty0
  (3):\penalty0 256--278, 1974.
\newblock ISSN 10902724.
\newblock \doi{10.1016/S0022-0000(74)80044-9}.

\bibitem[Kapuscinski and Tayur(1999)]{Kapuscinski1999OptimalSystems}
R.~Kapuscinski and S.~Tayur.
\newblock {Optimal policies and simulation-based optimization for capacitated
  production inventory systems}.
\newblock In \emph{Quantitative Models for Supply Chain Management}, pages
  7--40. Springer, 1999.

\bibitem[Kellerer et~al.(2004)Kellerer, Pferschy, and Pisinger]{Kellerer2004}
H.~Kellerer, U.~Pferschy, and D.~Pisinger.
\newblock \emph{{Knapsack Problems}}, volume~53.
\newblock Springer Verlag, Heidelberg, 2004.
\newblock ISBN 9788578110796.
\newblock \doi{10.1017/CBO9781107415324.004}.

\bibitem[Kong et~al.(2019)Kong, Sivakumar, Liaw, and Mehta]{Kong2019}
W.~Kong, D.~Sivakumar, C.~Liaw, and A.~Mehta.
\newblock {A new dog learns old tricks: RL finds Classic optimization
  algorithms}.
\newblock \emph{7th International Conference on Learning Representations, ICLR
  2019}, \penalty0 (2009):\penalty0 1--25, 2019.

\bibitem[Konno and Yamazaki(1991)]{Konno1991}
H.~Konno and H.~Yamazaki.
\newblock {Mean-Absolute Deviation Portfolio Optimization Model and Its
  Applications to Tokyo Stock Market}.
\newblock \emph{Management Science}, 37\penalty0 (5), 1991.

\bibitem[Kool et~al.(2019)Kool, Van~Hoof, and Welling]{Kool2019}
W.~Kool, H.~Van~Hoof, and M.~Welling.
\newblock {Attention, learn to solve routing problems!}
\newblock \emph{7th International Conference on Learning Representations, ICLR
  2019}, pages 1--25, 2019.

\bibitem[Krokhmal et~al.(2003)Krokhmal, Palmquist, and Uryasev]{Krokhmal2003}
P.~Krokhmal, J.~Palmquist, and S.~Uryasev.
\newblock {Portfolio optimization with conditional value-at-risk objective and
  constraints}.
\newblock \emph{Journal of Risk}, 4\penalty0 (2), 2003.

\bibitem[Lee et~al.(1997)Lee, Padmanabhan, and Whang]{Lee1997InformationEffect}
H.~L. Lee, V.~Padmanabhan, and S.~Whang.
\newblock {Information distortion in a supply chain: The bullwhip effect}.
\newblock \emph{Management Science}, 43\penalty0 (4):\penalty0 546--558, 4
  1997.
\newblock ISSN 00251909.
\newblock \doi{10.1287/mnsc.43.4.546}.
\newblock URL
  \url{https://pubsonline.informs.org/doi/abs/10.1287/mnsc.43.4.546}.

\bibitem[Li(2017)]{Li2017}
Y.~Li.
\newblock {Deep Reinforcement Learning: An Overview}.
\newblock pages 1--70, 2017.
\newblock ISSN 1701.07274.
\newblock \doi{10.1007/978-3-319-56991-8{\_}32}.
\newblock URL \url{http://arxiv.org/abs/1701.07274}.

\bibitem[Lueker(1995)]{Lueker1995}
G.~S. Lueker.
\newblock {Average-case analysis of off-line and on-line knapsack problems}.
\newblock \emph{Proceedings of the Annual ACM-SIAM Symposium on Discrete
  Algorithms}, \penalty0 (January 1995):\penalty0 179--188, 1995.

\bibitem[Ma et~al.(2019)Ma, Simchi-Levi, and Zhao]{Ma2019}
W.~Ma, D.~Simchi-Levi, and J.~Zhao.
\newblock {A Competitive Analysis of Online Knapsack Problems with Unit
  Density}.
\newblock \emph{SSRN Electronic Journal}, 2019.
\newblock \doi{10.2139/ssrn.3423199}.

\bibitem[Marchetti-Spaccamela and Vercellis(1995)]{Marchetti-Spaccamela1995}
A.~Marchetti-Spaccamela and C.~Vercellis.
\newblock {Stochastic on-line knapsack problems}.
\newblock \emph{Mathematical Programming}, 68\penalty0 (1-3):\penalty0 73--104,
  1995.
\newblock ISSN 14364646.
\newblock \doi{10.1007/BF01585758}.

\bibitem[Markowitz(1952)]{Markowitz1952}
H.~Markowitz.
\newblock {PORTFOLIO SELECTION}.
\newblock \emph{The Journal of Finance}, 7\penalty0 (1):\penalty0 77--91, 3
  1952.
\newblock ISSN 15406261.
\newblock \doi{10.1111/j.1540-6261.1952.tb01525.x}.
\newblock URL \url{http://doi.wiley.com/10.1111/j.1540-6261.1952.tb01525.x}.

\bibitem[Martinez et~al.(2011)Martinez, Nowe, Suarez, and Bello]{Martinez2011}
Y.~Martinez, A.~Nowe, J.~Suarez, and R.~Bello.
\newblock {A Reinforcement Learning Approach for the Flexible Job Shop
  Scheduling Problem}.
\newblock In C.~A. Coello, editor, \emph{Learning and Intelligent
  Optimization}, pages 253--262. Springer Verlag, 2011.

\bibitem[Mathews(1896)]{Mathews1896}
G.~B. Mathews.
\newblock {On the Partition of Numbers}.
\newblock \emph{Proceedings of the London Mathematical Society}, s1-28\penalty0
  (1):\penalty0 486--490, 11 1896.
\newblock ISSN 0024-6115.
\newblock \doi{10.1112/plms/s1-28.1.486}.
\newblock URL \url{https://doi.org/10.1112/plms/s1-28.1.486}.

\bibitem[Moritz et~al.(2018)Moritz, Nishihara, Wang, Tumanov, Liaw, Liang,
  Elibol, Yang, Paul, Jordan, and Stoica]{Moritz2018}
P.~Moritz, R.~Nishihara, S.~Wang, A.~Tumanov, R.~Liaw, E.~Liang, M.~Elibol,
  Z.~Yang, W.~Paul, M.~I. Jordan, and I.~Stoica.
\newblock {Ray: A Distributed Framework for Emerging AI Applications}.
\newblock 2018.
\newblock URL \url{http://arxiv.org/abs/1712.05889}.

\bibitem[Mortazavi et~al.(2015)Mortazavi, Arshadi~Khamseh, and
  Azimi]{Mortazavi2015}
A.~Mortazavi, A.~Arshadi~Khamseh, and P.~Azimi.
\newblock {Designing of an intelligent self-adaptive model for supply chain
  ordering management system}.
\newblock \emph{Engineering Applications of Artificial Intelligence},
  37:\penalty0 207--220, 2015.
\newblock ISSN 09521976.
\newblock \doi{10.1016/j.engappai.2014.09.004}.
\newblock URL \url{http://dx.doi.org/10.1016/j.engappai.2014.09.004}.

\bibitem[Nocedal and Wright(2006)]{Nocedal2006NumericalOptimization}
J.~Nocedal and S.~Wright.
\newblock \emph{{Numerical optimization}}.
\newblock Springer Science {\&} Business Media, 2006.

\bibitem[Oroojlooyjadid et~al.(2017)Oroojlooyjadid, Nazari, Snyder, and
  Tak{\'{a}}{\v{c}}]{Oroojlooyjadid2017}
A.~Oroojlooyjadid, M.~Nazari, L.~Snyder, and M.~Tak{\'{a}}{\v{c}}.
\newblock {A Deep Q-Network for the Beer Game: A Reinforcement Learning
  algorithm to Solve Inventory Optimization Problems}.
\newblock pages 1--38, 2017.
\newblock URL \url{http://arxiv.org/abs/1708.05924}.

\bibitem[Perold(1984)]{Perold1984}
A.~F. Perold.
\newblock {Large-scale portfolio optimization}.
\newblock \emph{Management Science}, 30\penalty0 (10), 1984.

\bibitem[Powell(1964)]{Powell1964AnDerivatives}
M.~J.~D. Powell.
\newblock {An efficient method for finding the minimum of a function of several
  variables without calculating derivatives}.
\newblock \emph{The Computer Journal}, 7\penalty0 (2):\penalty0 155--162, 1964.

\bibitem[Ram(1992)]{Ram1992}
B.~Ram.
\newblock {The pallet loading problem: A survey}.
\newblock \emph{International Journal of Production Economics}, 28\penalty0
  (2):\penalty0 217--225, 1992.
\newblock ISSN 09255273.
\newblock \doi{10.1016/0925-5273(92)90034-5}.

\bibitem[Sahinidis(2019)]{BARON2019}
N.~V. Sahinidis.
\newblock {BARON 19.7.13: Global Optimization of Mixed-Integer Nonlinear
  Programs}, 2019.
\newblock URL \url{http://www.minlp.com/downloads/docs/baron manual.pdf}.

\bibitem[Schulman et~al.(2016)Schulman, Moritz, Levine, Jordan, and
  Abbeel]{Schulman2016}
J.~Schulman, P.~Moritz, S.~Levine, M.~I. Jordan, and P.~Abbeel.
\newblock {High-dimensional continuous control using generalized advantage
  estimation}.
\newblock pages 1--14, 2016.
\newblock URL \url{https://arxiv.org/pdf/1506.02438.pdf}.

\bibitem[Silver et~al.(2017)Silver, Hubert, Schrittwieser, Antonoglou, Lai,
  Guez, Lanctot, Sifre, Kumaran, Graepel, Lillicrap, Simonyan, and
  Hassabis]{Silver2017}
D.~Silver, T.~Hubert, J.~Schrittwieser, I.~Antonoglou, M.~Lai, A.~Guez,
  M.~Lanctot, L.~Sifre, D.~Kumaran, T.~Graepel, T.~Lillicrap, K.~Simonyan, and
  D.~Hassabis.
\newblock {Mastering Chess and Shogi by Self-Play with a General Reinforcement
  Learning Algorithm}.
\newblock pages 1--19, 2017.
\newblock URL \url{http://arxiv.org/abs/1712.01815}.

\bibitem[Simchi-Levi and Zhao(2012)]{Simchi-Levi2012PerformanceSurvey}
D.~Simchi-Levi and Y.~Zhao.
\newblock {Performance Evaluation of Stochastic Multi-Echelon Inventory
  Systems: A Survey}.
\newblock \emph{Advances in Operations Research}, 2012:\penalty0 126254, 2012.
\newblock ISSN 1687-9147.
\newblock \doi{10.1155/2012/126254}.
\newblock URL \url{https://doi.org/10.1155/2012/126254}.

\bibitem[Song et~al.(2014)Song, Xiao, Chen, and Luo]{Song2014}
W.~Song, Z.~Xiao, Q.~Chen, and H.~Luo.
\newblock {Adaptive resource provisioning for the cloud using online bin
  packing}.
\newblock \emph{IEEE Transactions on Computers}, 63\penalty0 (11):\penalty0
  2647--2660, 2014.
\newblock ISSN 15579956.
\newblock \doi{10.1109/TC.2013.148}.

\bibitem[Sutton and Barto(2018)]{Sutton2018}
R.~Sutton and A.~Barto.
\newblock \emph{{Reinforcement Learning: An Introduction}}.
\newblock MIT Press, Cambridge, Massachusetts, 2 edition, 2018.
\newblock URL \url{http://incompleteideas.net/book/bookdraft2017nov5.pdf}.

\end{thebibliography}
\newpage
\appendix
\section{Model Sizes}
\label{sec:model_sizes}

\begin{table}[!ht]
\centering
\caption{Average model sizes and solution times for MILP models.}
\label{tab:model_sizes}
\resizebox{\columnwidth}{!}{%
\begin{tabular}{@{}lcccccc@{}}\toprule
Model & Number of Variables & Number of Constraints & Time to Solve (s) \\
\midrule
BinKp MIP & 201   & 2     & 0.023 \\
BKP MIP & 201   & 202   & 0.01  \\
OKP MIP & 51    & 3     & 0.006 \\
VM Packing MIP & 134,235  & 143,901 & 1.38  \\
InvManagement MIP & 1,000 & 1,387 & 8.149 \\
InvManagement Oracle & 541   & 631   & 0.009 \\
InvManagement oMIP  & 967   & 1,341 & 8.248 \\
Det MPAA & 101   & 41    & 0.017 \\
RO MPAA & 102   & 42    & 503.5 \\
\bottomrule
\end{tabular}%
}
\end{table}

\end{document}